\documentclass[letterpaper]{article} 
\usepackage{aaai2027}  
\usepackage[hyphens]{url}  
\usepackage{graphicx} 
\usepackage{natbib}  
\usepackage{caption} 
\usepackage{booktabs}
\usepackage{multirow}
\usepackage{arydshln}
\usepackage{listings}
\usepackage{listings}
\usepackage{xcolor}
\usepackage{amsmath}
\usepackage{algorithm}
\usepackage{algpseudocode}

\definecolor{codegray}{RGB}{245,245,245}

\lstdefinestyle{pythonstyle}{
  language=Python,
  basicstyle=\ttfamily\tiny,
  keywordstyle=\color{blue},
  stringstyle=\color{orange},
  commentstyle=\color{green!60!black},
  numbers=left,
  numberstyle=\tiny\color{gray},
  stepnumber=1,
  numbersep=5pt,
  showstringspaces=false,
  tabsize=4,
  breaklines=true,
  backgroundcolor=\color{codegray},
  frame=none,
  linewidth=0.95\linewidth,       
  xleftmargin=0.05\linewidth,    
  xrightmargin=0.025\linewidth    
}

\lstdefinestyle{markdownstyle}{
  language=,
  basicstyle=\ttfamily\tiny,
  numbers=none,
  backgroundcolor=\color{codegray},
  frame=none,
  linewidth=0.95\linewidth,
  xleftmargin=0.025\linewidth,
  xrightmargin=0.025\linewidth,
  breaklines=true,           
  breakatwhitespace=true,    
  postbreak=\mbox{},         
  breakindent=0pt            
}

\usepackage{newfloat}
\usepackage{listings}
\DeclareCaptionStyle{ruled}{labelfont=normalfont,labelsep=colon,strut=off} 
\floatstyle{ruled}
\newfloat{listing}{tb}{lst}{}
\floatname{listing}{Listing}

\usepackage{booktabs}
\usepackage{multirow}
\usepackage[table]{xcolor}

\definecolor{IDGray}{gray}{0.92}

\usepackage{booktabs}

\usepackage{amsmath,amssymb,bm}
\usepackage{xcolor}
\usepackage{tikz}
\usetikzlibrary{arrows.meta,positioning,fit}

\definecolor{adblue}{HTML}{0072B2}
\definecolor{adorange}{HTML}{E69F00}
\definecolor{adgreen}{HTML}{009E73}
\definecolor{adgray}{HTML}{5B6573}

\title{Analytic Dynamics: Learning Physics-Grounded Representation for Fast Intrinsic Dynamics Inference from Monocular Videos}
\author {
    Jiajing Lin\equalcontrib,
    Jikuan Zhange\equalcontrib,
    Jianhua Sun\corresponding
}
\affiliations {
    School of Artificial Intelligence, Shanghai Jiao Tong University\\
    \{jiajinglin, k-u-a-n\_as, gothic\}@sjtu.edu.cn
}

\nocopyright

\begin{document}

\maketitle

\begin{abstract}
Inferring object dynamics from visual observations is essential for intelligent agents to reason about and interact with the physical world, yet remains challenging due to the fundamental gap between visual evidence and intrinsic dynamics. Existing methods either rely on costly per-scene optimization, limiting efficiency and scalability, or directly map visual evidence to intrinsic dynamics without intermediate physical abstractions, making them prone to appearance and geometry shortcuts. To bridge this gap, we propose Analytic Dynamics, a feed-forward dynamics inference framework that introduces an intermediate physics-grounded dynamics representation between visual observations and intrinsic dynamics. Specifically, we leverage privileged physical states, including position, displacement, and deformation gradient fields, which are available in simulation, to learn a structured dynamics representation that is difficult to discover from visual observations alone. By aligning visual representations with this space, we equip visual models with a physics-grounded inductive bias, guiding them to capture dynamics-relevant patterns for material model classification and parameter regression. To facilitate this research, we develop a dynamics data generation pipeline and benchmark containing paired physical state trajectories, rendered videos, and ground-truth material models and parameters. Extensive experiments demonstrate that Analytic Dynamics achieves efficient, accurate, and generalizable dynamics inference from monocular videos.

\end{abstract}


\begin{figure*}[!t]
	\centering
	\includegraphics[width=1\linewidth]{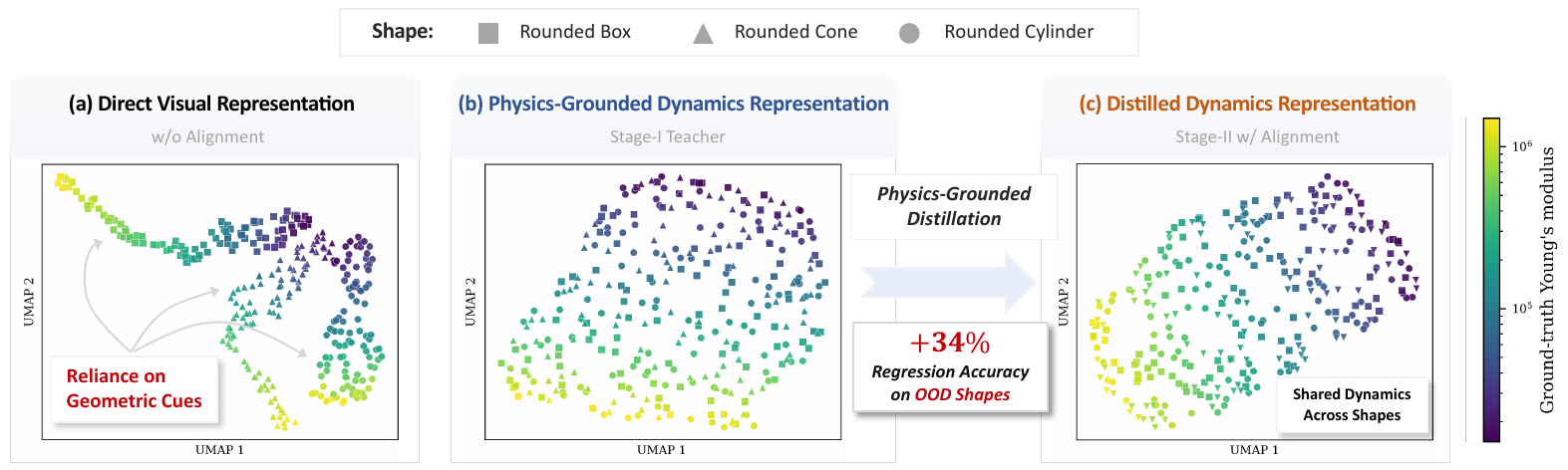}
	\caption{
    \textbf{Physics-grounded distillation reduces shape dependence.}
    We use UMAP to visualize representations of elastic samples from three training shapes: Box, Cone, and Cylinder. Marker types denote object shapes, while colors indicate the ground-truth Young's modulus $E$. Without alignment, direct visual representations form three shape-specific branches, revealing strong reliance on geometric cues. In contrast, the Stage I teacher organizes samples into a smooth continuum according to $E$, forming a shared dynamics structure across geometries. Distillation transfers this organization to Stage II, enabling the visual model to capture dynamics-relevant patterns across shapes and improving $E$ regression accuracy by 34\% on unseen shapes.
    }
    \label{fig:intro}
\end{figure*}

\section{Introduction}


Embodied agents~\cite{simulatorembodied, worldmdoelsurvey} need to reason about how objects respond to physical interactions to make reliable decisions. This capability relies on understanding the intrinsic dynamics of objects, including the underlying material models and material parameters. Humans effortlessly infer such dynamics from observed motions, enabling them to anticipate future behaviors and interact effectively with the physical world. Endowing machines with similar intuitive physics~\cite{intuitivephysics} is a fundamental goal of embodied intelligence~\cite{sun2025digital}.

With advances in differentiable rendering~\cite{NeRF,3DGS} and simulation~\cite{jiang2016mpm}, a line of work~\cite {pacnerf, gic, visionlaw} formulates dynamics inference as inverse physics problems, optimizing material parameters to match simulated behaviors with multi-view videos. While achieving accurate estimation, these methods require extensive simulation–optimization iterations from scratch for each scene, resulting in high computational cost and limited scalability to unseen objects.
Alternatively, another line of work~\cite {nerf2physics, pugs, physgs} explores prompting large vision-language models to reason about intrinsic dynamics from visual observations. While offering efficient and versatile inference, these approaches primarily rely on commonsense knowledge acquired from large-scale corpora, resulting in noisy and coarse-grained predictions.

To explore a better balance between inference efficiency and physical accuracy, recent works~\cite{pixie, unipixie, slatphys} leverage large-scale synthetic 3D asset–dynamics pairs to train feed-forward models. However, static 3D assets only provide appearance and geometry, which are inherently ambiguous for inferring intrinsic dynamics: visually identical objects may correspond to different material models or parameters.
In contrast, monocular videos capture how objects move and deform under physical interactions, providing behavioral evidence induced by their intrinsic dynamics.
However, inferring intrinsic dynamics from monocular videos remains challenging, as videos only capture 2D projections of physical responses induced by intrinsic dynamics, which are mixed with factors unrelated to dynamics, such as viewpoint, geometry, and appearance.
Without explicit physical guidance, models can easily exploit spurious correlations rather than learning dynamics-relevant patterns, leading to poor generalization.

To address the aforementioned challenges, we propose~{\em Analytic Dynamics}, a feed-forward framework for efficiently inferring constitutive material models and parameters from monocular videos. Our key insight is to introduce a physics-grounded dynamics representation between visual observations and intrinsic dynamics, providing visual models with a physically meaningful learning target. Specifically, we develop a two-stage privileged dynamics distillation framework. First, we leverage privileged physical states as input, including position, displacement, and deformation gradient fields, which are only accessible in simulation, to learn a structured dynamics representation space for material model classification and parameter regression. Since these states directly describe physical responses governed by intrinsic dynamics, the learned representation naturally encodes dynamics-related information, resulting in a physics-grounded representation space. Second, we align monocular video with this learned dynamics space. This alignment introduces a physics-grounded inductive bias, guiding visual representations toward dynamics-related information reflected in object motion and deformation, and enabling more accurate and generalizable inference of material models and parameters.
To support this research, we build a simulator-based benchmark of 9,000 dynamics instances across 60 objects, covering diverse material models, parameters, and views. Each instance pairs monocular videos with physical-state trajectories and ground-truth material labels.
Our contributions are summarized as follows:
\begin{itemize}
    \item We propose a feed-forward dynamics inference framework that is the first to efficiently recover constitutive material models and parameters from monocular videos.
    \item We introduce a two-stage privileged dynamics distillation framework that learns a physics-grounded dynamics representation from physical states and transfers it to the visual model, encouraging the model to capture dynamics-relevant patterns for improved generalization.
    \item We develop a dynamics data generation pipeline and benchmark for intrinsic dynamics inference. Extensive experiments demonstrate that~{\em Analytic Dynamics} outperforms existing baselines, enabling efficient inference within 3.83 milliseconds while maintaining strong out-of-distribution generalization.
\end{itemize}


\begin{figure*}[!t]
    \centering
    \includegraphics[width=\linewidth]{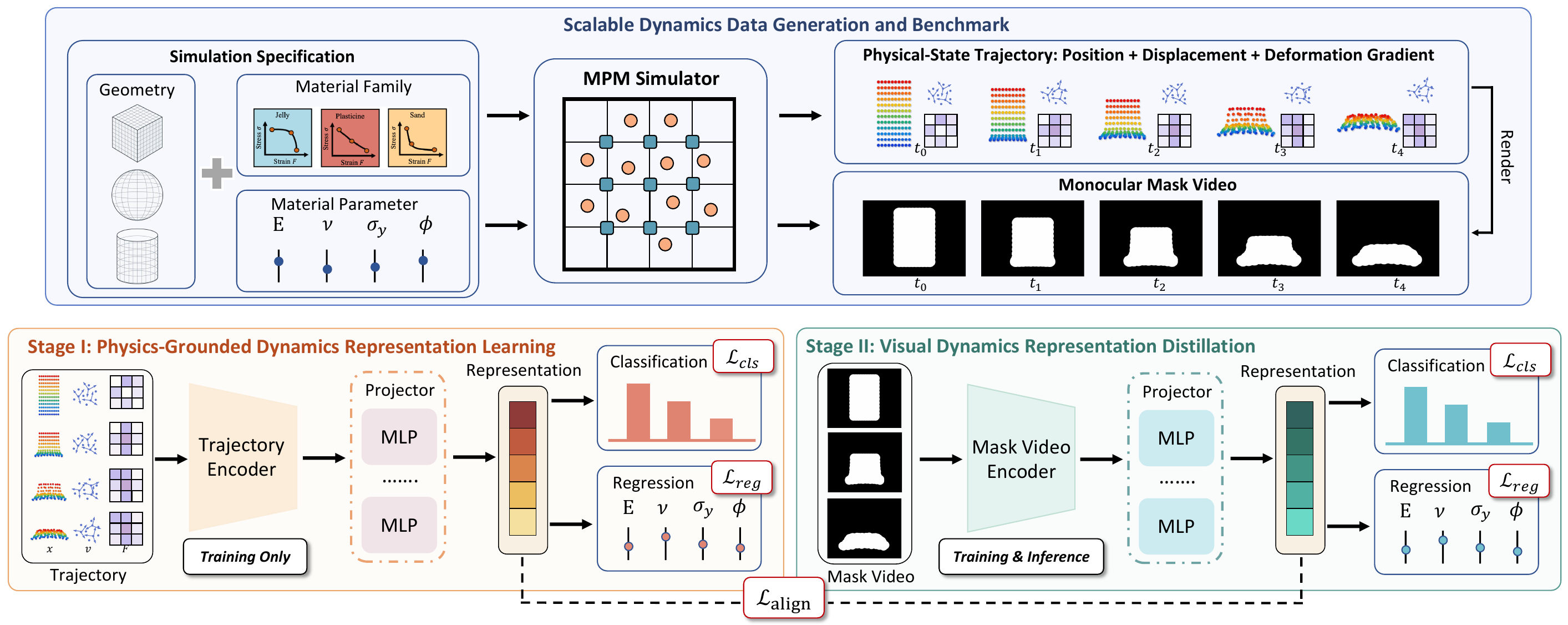}
    \caption{\textbf{Overview of~{\em Analytic Dynamics}.}
Given sampled geometry, material family, material parameters, and external
excitation, an MPM simulator generates paired tracked particle trajectories
and synchronized monocular mask videos. The trajectories contain particle
positions, displacements, and deformation gradients. In Stage~I, a trajectory
teacher learns a dynamics-relevant representation from these simulator-only
states using classification and parameter-regression supervision. 
In Stage~II, a video student learns the same inference task from mask videos while aligning its representation with the frozen teacher. 
At test time, only the video student is retained to predict intrinsic dynamics from readily captured monocular videos.
}
    \label{fig:pipeline}
\end{figure*}

\section{Related Work}
\subsection{Per-Scene Dynamics Optimization}
Per-scene inverse-physics methods identify object dynamics from videos by optimizing physical parameters through differentiable rendering and simulation. PAC-NeRF, GIC, Spring-Gaus, and M-PhyGs~\cite{pacnerf, gic, springgaus,mphygs} operate under prescribed material models, progressively supporting explicit geometry, heterogeneous elasticity, and multi-material objects. NeuMA and MASIV~\cite{neuma,masiv} relax fixed constitutive assumptions through neural constitutive modeling, whereas VisionLaw~\cite{visionlaw} searches for interpretable symbolic laws. PhysDreamer, DreamPhysics, Physics3D, PhysFlow, and OmniPhysGS~\cite{physdreamer,dreamphysics,physics3d,physflow,omniphysgs} instead use video diffusion priors to optimize physical properties without captured motion. Vid2Sim combines feed-forward prediction with lightweight per-scene refinement \cite{vid2sim}. Despite these advances, optimization-based methods remain computationally expensive and require scene-specific fitting, limiting their scalability.
\subsection{Feed-Forward Dynamics Inference}
To avoid expensive scene-specific system identification, feed-forward methods directly predict physical properties from visual observations. NeRF2Physics, GaussianProperty, PUGS, and PhysGS~\cite{nerf2physics,gaussianproperty, pugs} transfer material semantics and physical priors from pretrained vision-language models (VLMs) to 3D scene representations for zero-shot property estimation; among them, PUGS improves property propagation over 3D Gaussians, while PhysGS~\cite{physgs} explicitly models prediction uncertainty. Phys4DGen and PhysSplat~\cite{phys4dgen,physsplat} extend this strategy to physics-driven 4D generation through explicit simulation. While efficient and broadly applicable, these methods are constrained by VLM commonsense priors and may produce coarse estimates. Another line, exemplified by PIXIE~\cite{pixie}, learns feed-forward material inference from large-scale annotated 3D assets. Building on this paradigm, SLAT-Phys enables efficient single-image prediction, while UniPixie~\cite{slatphys,unipixie} extends deterministic estimation to controllable material distributions. However, these methods may exploit appearance and geometry shortcuts rather than learn generalizable physical regularities.

\section{Methodology}
\label{sec:guided_method}

\subsection{Problem Formulation and Overview}
\label{sec:guided_overview}


For a dynamics instance $i$, let $\mathcal{V}_i^{(r)}$
denote a monocular mask video of a moving object observed
from viewpoint $r$. Our objective is to learn a feed-forward
video model $\mathcal{M}_{\Phi}$ that predicts the material
family $y_i\in\mathcal{C}$ and its corresponding material
parameters $\boldsymbol{\theta}_i$:
\begin{equation}
    \left(
        \hat{y}_{i},
        \hat{\boldsymbol{\theta}}_{i}
    \right)
    =
    \mathcal{M}_{\Phi}
    \left(
        \mathcal{V}_i^{(r)}
    \right).
    \label{eq:guided_inference}
\end{equation}

Directly learning this mapping from videos is challenging, as a monocular video provides only an indirect 2D observation of the physical response. In simulation, each dynamics instance additionally provides a tracked physical-state trajectory $\mathcal{S}_i$, offering a more direct observation of the underlying physical response.
We therefore use these training-only physical states to guide the learning of $\mathcal{M}_{\Phi}$, while relying only on video inputs at inference.

To this end, we propose a two-stage privileged dynamics distillation framework. We first develop a data generation pipeline that produces paired physical-state trajectories and monocular mask videos for each simulated dynamics instance. In Stage I, a trajectory teacher learns a physics-grounded representation predictive of the material family and its parameters. In Stage II, this representation is distilled into a video student through representation alignment and direct task supervision. The overall framework of \emph{Analytic Dynamics} is illustrated in Fig.~\ref{fig:pipeline}.

\subsection{Dynamics Data Generation and Benchmark}
\label{sec:guided_generation}
To support privileged dynamics distillation, we construct paired physical-state trajectories and monocular videos from the same simulated dynamics instance. Each instance is defined by an object geometry, a material family, material parameters, and an external excitation.
Given these configurations, we use the Material Point Method (MPM)~\cite{jiang2016mpm,ma2023nclaw} to simulate the object response, producing temporally evolving particle states together with a synchronized deforming surface.

To construct the physical-state trajectory, we perform farthest-point sampling (FPS) only on the initial particle set and track the same particles throughout the simulation. 
This preserves particle identity over time and converts independent state snapshots into Lagrangian trajectories. 
For the $k$-th tracked particle, we compute its frame displacement as
\begin{equation}
    \Delta \mathbf{x}_{i,t,k} =
\begin{cases} 
0, & t = 0, \\
\mathbf{x}_{i,t,k} - \mathbf{x}_{i,t-1,k}, & t \geq 1.
\end{cases}
\end{equation}
To encourage the model to focus on relative motion patterns rather than absolute displacement magnitudes, we normalize the displacements using their median magnitude over all tracked particles and non-initial frames: 
\begin{align}
s_i = &\text{median}_{t=1, \dots, T-1} \left( \left\| \Delta \mathbf{x}_{i,t,k} \right\|_2 \right), \\
&{\mathbf{v}}_{i,t,k} = \frac{\Delta \mathbf{x}_{i,t,k}}{s_i}.
\end{align}
The resulting physical-state trajectory $\mathcal{S}_i$ consists of tracked particle positions $\mathbf{x}_i$, normalized frame displacements $\mathbf{v}_i$, and deformation gradients $\mathbf{F}_i$.

To generate the corresponding visual observations, we bind each surface-mesh vertex to nearby material particles in the initial configuration. During simulation, the mesh vertices are advected by interpolating the motion of their associated particles, yielding a deforming surface synchronized with the particle trajectory. We render this surface from multiple viewpoints into a monocular mask video $\{\mathcal{V}_i^r\}_{r=1}^R$. 

Each simulated instance therefore provides one physical-state trajectory, multiple synchronized monocular videos, and their shared material annotations:
\begin{equation}
    \mathcal{D}_i = \left( \mathcal{S}_i, \{ \mathcal{V}_i^{(r)} \}_{r=1}^R, y_i, \theta_i \right).
\end{equation}
The trajectory is used to train the Stage-I teacher, while the paired videos support Stage-II representation distillation and task supervision.

\subsection{Privileged Dynamics Representation Learning}
\label{sec:guided_teacher}
Stage I aims to learn a teacher representation from privileged physical-state trajectories. Unlike monocular videos, these trajectories directly capture 3D particle motion and deformation, encouraging the model to learn dynamics-relevant patterns.

Given a physical-state trajectory $\mathcal{S}_i$, P4Transformer~\cite{fan2021p4transformer} models local and long-range spatiotemporal interactions among the tracked particles. A projector then maps the pooled trajectory feature to a teacher representation $\mathbf{z}_i^T \in \mathbb{R}^d$, which defines the representation space used for subsequent visual distillation.

To make this representation predictive of intrinsic dynamics, we train the teacher with material-family classification and parameter regression. Classification distinguishes discrete material families, while regression preserves continuous parameter variations within each family. 
Since different material families involve distinct parameter sets and numerical scales, we normalize each parameter and compute RegL1 as the mean L1 error over the parameters defined for the ground-truth family:
\begin{equation}
\label{eq:RegL1}
\mathcal{L}_{\mathrm{RegL1}}=
    \frac{
\sum_{i=1}^{B}\sum_{j=1}^{d_{\max}}
m_{i,j}
\left|
\hat{\overline{\theta}}_{i,j}
-
\overline{\theta}_{i,j}
\right|
}{
\sum_{i=1}^{B}\sum_{j=1}^{d_{\max}}m_{i,j}
}.
\end{equation}
Here, $\overline{\theta}_{i,j}$ and $\hat{\overline{\theta}}_{i,j}$ denote the normalized ground-truth and predicted parameters, respectively, and $m_{i,j}$ indicates whether the $j$-th parameter is defined for sample $i$. Parameter normalization places different physical quantities on a comparable scale, while the validity mask enables unified optimization across material families. 
The classification branch is trained with standard cross-entropy $\mathcal{L}_{cls}$. 
The overall Stage-I objective is:
\begin{equation}
    \mathcal{L}_{\mathrm{Stage-I}} = \lambda_{cls}\mathcal{L}_{cls} + \lambda_{reg}\mathcal{L}_{\mathrm{RegL1}}
\end{equation}

By combining direct 3D physical observations with material supervision, Stage I learns a physics-grounded representation predictive of the material family and its parameters.
After training, the trajectory encoder and projector are frozen, and $\mathbf{z}^T$ serves as the representation target for the paired monocular videos in Stage II.

\subsection{Visual Dynamics Representation Distillation}
\label{sec:guided_distillation}
Stage II distills the physics-grounded representation learned in Stage I into a deployable video model. 
Given a monocular mask video $\mathcal{V}_i^{(r)}$, a UniFormer encoder~\cite{li2022uniformer} followed by a projector produces a visual representation $\mathbf{z}_{i,r}^{V}\in\mathbb{R}^{d}$. 
We use mask videos to retain silhouette motion and deformation while removing variations in texture and color.
Meanwhile, the frozen teacher extracts the target representation $\mathbf{z}_i^{T}$ from the paired physical-state trajectory $\mathcal{S}_i$. Different views of the same dynamics instance share the same teacher target, encouraging view-consistent visual representations.

We align the visual and teacher representations using cosine and $\ell_1$ distances:
\begin{equation}
\mathcal{L}_{\mathrm{align}}
=
\frac{1}{B}
\sum_{i=1}^{B}
\left[
1-
\left(
\widetilde{\mathbf{z}}_{i,r}^{V}
\right)^{\top}
\widetilde{\mathbf{z}}_{i}^{T}
+
\alpha
\left\|
\mathbf{z}_{i,r}^{V}
-
\mathbf{z}_{i}^{T}
\right\|_{1}
\right],
\label{eq:alignment_loss}
\end{equation}
where
$\widetilde{\mathbf{z}}
=
\mathbf{z}/(\|\mathbf{z}\|_2+\varepsilon)$,
and $\alpha$ balances the two terms. The cosine term aligns the overall feature direction, while the $\ell_1$ term reduces their element-wise discrepancy.
The teacher remains frozen throughout Stage II, so the alignment loss updates only the video branch.
The video model uses the same material-family classification and parameter-regression design as the teacher, supervised by cross-entropy and RegL1, respectively. The overall Stage-II objective is
\begin{equation}
\mathcal{L}_{\mathrm{Stage\text{-}II}}
=
\lambda_{\mathrm{align}}
\mathcal{L}_{\mathrm{align}}
+
\lambda_{\mathrm{cls}}
\mathcal{L}_{\mathrm{cls}}
+
\lambda_{\mathrm{reg}}
\mathcal{L}_{\mathrm{RegL1}}.
\label{eq:stage2_loss}
\end{equation}

Representation alignment guides the video model toward features learned from direct 3D motion and deformation, while task supervision keeps the visual representation predictive of material families and parameters. At inference, the teacher is discarded, and the video model directly predicts intrinsic dynamics from a monocular mask video.

\begin{table*}[t]
\centering
\begin{tabular}{c|ccc|cccc|cc}
\toprule
\multirow{2}{*}{Method} 
& \multicolumn{3}{c|}{Jelly}
& \multicolumn{4}{c|}{Plasticine}
& \multicolumn{2}{c}{Sand} \\

\cmidrule(lr){2-4}
\cmidrule(lr){5-8}
\cmidrule(lr){9-10}

& $Acc.\uparrow$
& $E\downarrow$
& $\nu\downarrow$

& $Acc.\uparrow$
& $E\downarrow$
& $\nu\downarrow$
& $Yield\downarrow$

& $Acc.\uparrow$
& $\phi\downarrow$ \\

\midrule

Per-scene Optimization
& - & 0.179 & 0.104 & - & 0.752 & 0.101 & \textbf{0.065} & - & 3.209
\\

\midrule

VLM baseline 
& 0.633 & 2.163 & 0.134 & 0.367 & 2.007 & 0.052 & 1.115 & 0.000 & -
\\

Video-LLM baseline 
& 0.847 & 1.376 & 0.118 & 0.833 & 0.756 & 0.045 & 0.785 & 0.327 & 7.803
\\

\noalign{\vskip 2pt\global\arrayrulewidth=0.6pt}
\cdashline{1-10}[3pt/2pt]
\noalign{\global\arrayrulewidth=0.4pt\vskip 3.5pt}

Ours (w/o alignment)
& 0.997 & 0.265 & 0.043 & 1.000 & 0.223 & 0.024 & 0.115 & 0.943 & 1.991
\\

Ours (alignment only)
& 1.000 & 0.188 & \textbf{0.036} & 1.000 & 0.146 & 0.021 & 0.106 & \textbf{0.973} & 1.927
\\

Ours (Full)
& \textbf{1.000} & \textbf{0.175} & 0.037 & \textbf{1.000} & \textbf{0.141} & \textbf{0.021} & 0.101 & 0.970 & \textbf{1.777}
\\

\bottomrule
\end{tabular}
\caption{\textbf{Quantitative comparison on objects unseen during training.} We report material classification accuracy ($\uparrow$) and material-specific parameter errors ($\downarrow$) for elasticity, plasticity, and sand. Young's modulus $E$ and yield stress $\sigma_y$ are evaluated using log-MAE, while Poisson's ratio $\nu$ and friction angle $\phi$ are evaluated using MAE.}
\label{tab:comparison}
\end{table*}

\begin{figure*}[!t]
	\centering
	\includegraphics[width=1\linewidth]{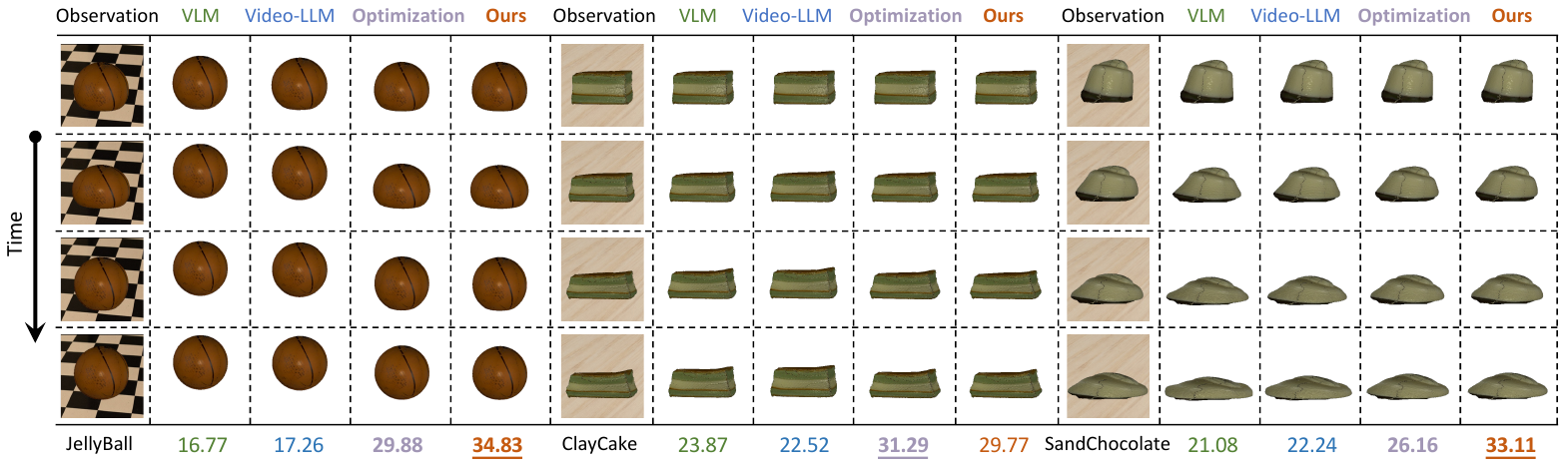}
	\caption{\textbf{Visual dynamics reconstruction on unseen objects.} We re-simulate each object using the intrinsic dynamics inferred by different methods. PSNR ($\uparrow$) between the reconstructed and observed frames is reported below each sequence.}
    \label{fig:visulization}
\end{figure*}

\section{Experiments}
\subsection{Experimental Setup}
\subsubsection{Implementation Details}
For the Shape-OOD experiments, the Stage-I teacher receives 25-frame
state trajectories of 1,024 tracked particles. Its input combines particle positions
\(\mathbf{x}\), displacement \(\mathbf{v}\), and deformation gradients
\(\mathbf{F}\). We use a P4Transformer~\cite{fan2021p4transformer} encoder with a 256-dimensional latent
space, four Transformer layers, and eight attention heads. The teacher is
trained for 300 epochs. In Stage~II, UniFormer-XXS~\cite{li2022uniformer} encodes 25-frame, \(128 \times 128\), single-channel mask videos. The complete Stage-I teacher is frozen, while the video encoder, projection head, classifier, and material-specific regression heads are optimized for 20 epochs with a batch size of 32.
Both stages are optimized using AdamW with an initial learning rate of
\(10^{-4}\), weight decay of \(10^{-4}\), and cosine decay to \(10^{-6}\). 
The Stage-II loss weights are \(\lambda_{\mathrm{cls}}=1.0\), \(\lambda_{\mathrm{reg}}=0.2\), and \(\lambda_{\mathrm{align}}=0.1\). 
All experiments are conducted on a single NVIDIA RTX 4090 GPU.

\subsubsection{Baselines}
We compare against three representative baselines spanning two paradigms.
First, following recent studies~\cite{nerf2physics, pugs} that leverage multimodal large language models for physical property reasoning, we construct a VLM baseline and a Video-LLM baseline.
They perform feed-forward inference of material models and parameters from an input image and video, respectively, using the material commonsense priors encoded in pretrained models.
The prompts are provided in the appendix.
Second, we consider per-scene inverse-physics optimization~\cite{gic, pacnerf}, which iteratively estimates the material parameters of an MPM simulator by matching the simulated dynamics to the observed video.
We adopt the implementation of VisionLaw~\cite{visionlaw} for this baseline.

\subsubsection{Datasets and Metrics}

We construct an MPM-based paired dynamics dataset comprising 9,000 dynamics instances across three material families: \texttt{jelly}, \texttt{plasticine}, and \texttt{sand}, with their constitutive formulations provided in the appendix. For \texttt{jelly}, we vary Young's modulus ($E$) and Poisson's ratio ($\nu$); for \texttt{plasticine}, we additionally vary the yield stress ($\sigma_y$); and for \texttt{sand}, we vary only the friction angle ($\phi$), the dominant material parameter governing its dynamics.
The benchmark covers 60 object shapes, including hand-crafted primitives and common real-world objects from OmniObject3D~\cite{omniobject3d}, evenly split into 30 training and 30 test shapes.
For each dynamics instance, we sample an object shape, material parameters, and external excitation, and generate synchronized particle-state trajectories and monocular mask videos from multiple views.
To mitigate viewpoint bias, we randomly sample a fixed number of views per instance during training.

We report material classification accuracy and material-specific parameter regression errors. Young's modulus \(E\) and yield stress \(\sigma_y\) are evaluated using log-MAE, while Poisson's ratio \(\nu\) and the friction angle \(\phi\) are evaluated using MAE. For an aggregate evaluation of parameter inference across material families and parameters, we use RegL1 described in Eq.~\ref{eq:RegL1}.
Lower RegL1 indicates more accurate parameter inference. 
Finally, we re-simulate each object using its inferred material parameters and report PSNR between the rendered reconstruction and the observed frames to assess visual dynamics
reconstruction quality.

\begin{figure*}[!t]
	\centering
	\includegraphics[width=1\linewidth]{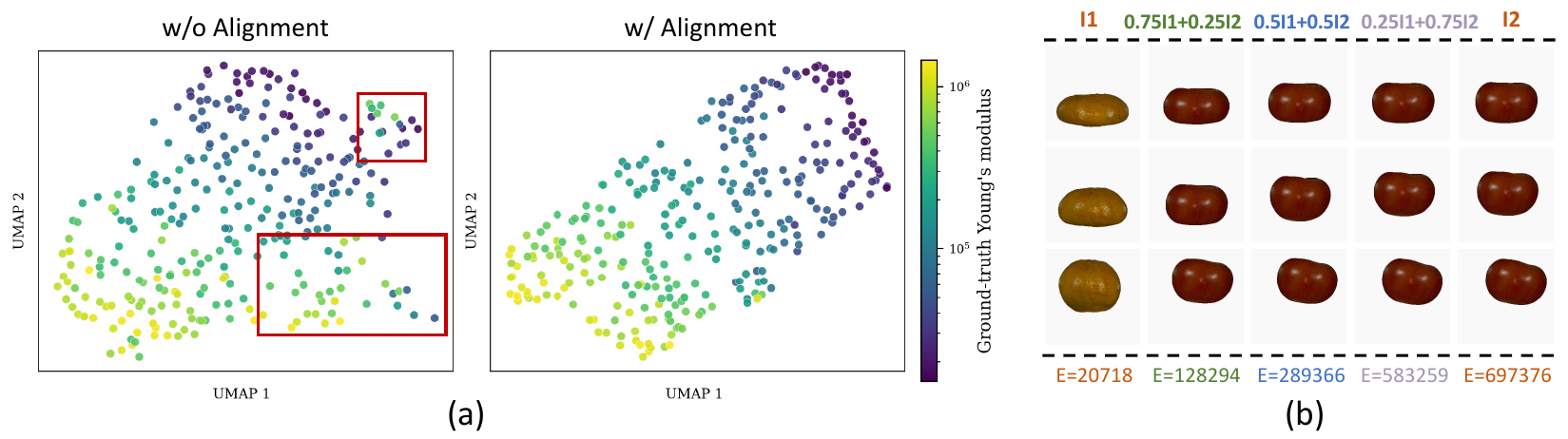}
	\caption{
    \textbf{Analysis of learned dynamics representations on unseen objects.}
    (a) UMAP visualizations of Stage II representations without and with alignment, colored by the ground-truth Young’s modulus.
    (b) Linear interpolation between two jelly representations from different objects with different stiffnesses. The top row shows the interpolation coefficients, and the bottom row shows the Young’s moduli decoded from the interpolated features.
    }
    \label{fig:representation analysis}
\end{figure*}

\subsection{Comparison on Unseen Objects}
We first evaluate material model classification and parameter regression on objects unseen during training. As shown in Tab.~\ref{tab:comparison}, the full model achieves the best overall performance. VLMs and Video-LLMs achieve reasonable classification accuracy but produce substantially larger errors in material parameters.
This suggests that vision-language models possess coarse-grained knowledge of materials and dynamics but struggle to precisely map visual dynamics to continuous constitutive parameters.
Despite its high computational cost, per-scene optimization remains competitive on several metrics. However, material parameters are only indirectly constrained by the final rendering loss, which entangles errors from physical simulation, 3D representation, and rendering. This makes it difficult to attribute visual discrepancies to material parameters. Under large deformations, the degradation and rendering artifacts of 3DGS further amplify this issue.

The results across model variants further validate privileged dynamics distillation. Alignment alone outperforms direct label supervision, indicating that the dynamics representation learned in Stage I provides the video model with a more physically meaningful learning target than material labels alone. Adding task supervision yields further gains: alignment captures dynamics-relevant information, while task supervision makes it predictive of material classes and parameters.

Fig.~\ref{fig:visulization} further examines whether the predicted parameters can reproduce the observed dynamics.
Specifically, we fix the object geometry and initial conditions, simulate the object using the material model and parameters predicted by each method, and compare the resulting sequence with the observed video.
The VLM and Video-LLM results are semantically plausible, but their deformation magnitudes and dynamic evolution deviate from the observations.
Our method more accurately reproduces the elastic deformation of JellyBall and the flow and accumulation of SandChocolate.
On ClayCake, per-scene optimization achieves higher reconstruction accuracy, while our method produces visually comparable deformation without any test-time optimization.

Overall, the quantitative results and visual dynamics reconstructions demonstrate that~{\em Analytic Dynamics} effectively infers material models and parameters from monocular videos in 3.83 ms and generalizes to unseen objects.

\subsection{Analysis of Learned Dynamics Representations}
To examine whether distillation reduces the dependence of visual representations on object geometry and organizes them by intrinsic dynamics, we visualize the representation space on unseen objects. If a model relies on geometric correlations in the training data, its structure should break under shape changes. In contrast, a dynamics-aware representation should preserve smooth neighborhoods organized by material parameters across different shapes.

As shown in Figure~\ref{fig:representation analysis}(a), the unaligned model exhibits local mixing and space folding: samples with similar Young’s modulus are scattered, while nearby points can have very different stiffness. Its representation is therefore not consistently organized by dynamics and remains entangled with dynamics-irrelevant factors such as shape and viewpoint, limiting cross-shape generalization. In contrast, the aligned representation varies more smoothly with Young’s modulus, and samples with similar stiffness remain close even across unseen objects. This indicates that Stage I distillation introduces a transferable dynamics structure and reduces reliance on object-specific geometric cues.

Figure~\ref{fig:representation analysis}(b) further evaluates whether the space continuously encodes material variation. We select two unseen jelly samples with different stiffness, linearly interpolate their dynamics representations, and feed them into the regression head. As the interpolation coefficient changes, the predicted \(E\) increases monotonically from $20718$ to $697376$, while the simulated deformation transitions from large to small. 
This shows that the learned space forms a continuous and ordered structure over material parameters. Overall, Stage I learns a representation organized by dynamics, and distillation effectively transfers this structure to the visual model.

\begin{figure}[!t]
	\centering
	\includegraphics[width=1\linewidth]{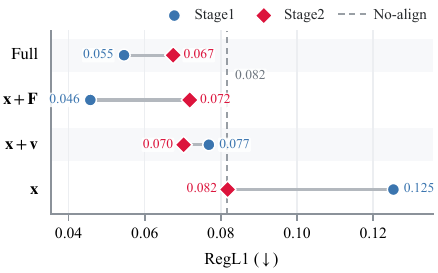}
	\caption{
    \textbf{Ablation of privileged physical states.}
    We compare position $x$, displacement $v$, and deformation gradient $F$ as inputs to the Stage I teacher, and report the RegL1 ($\downarrow$) errors of both the teacher and its corresponding Stage II student. Gray lines connect each teacher--student pair, while the dashed line denotes the Stage II baseline without distillation.
    }
    \label{fig:privileged ablation}
\end{figure}

\subsection{Effect of Privileged Physical States}
To study how privileged physical states affect representation learning and cross-modal distillation, we train Stage I teachers with position ($\mathbf{x}$), position and displacement ($\mathbf{v}$), position and displacement ($\mathbf{x} + \mathbf{v}$), position and deformation gradient ($\mathbf{x} + \mathbf{F}$) and the full state ($\mathbf{x} + \mathbf{v} + \mathbf{F}$), and distill each representation into the Stage II video model. Fig.~\ref{fig:privileged ablation} reports the RegL1 errors of parameter regression for both stages. 
The results show that dynamics discriminability does not always translate into visual transferability. 
Using position alone yields the worst Stage I performance. Position trajectories mainly preserve object geometry, making the learned representation more sensitive to shape than to dynamics-relevant information and thus less effective for guiding Stage II.
When the deformation gradient ($\mathbf{F}$) is added, the Stage I error drops significantly to 0.046, indicating that the deformation gradient provides strong cues for intrinsic dynamics. However, the most accurate teacher does not produce the best student. Although ($\mathbf{x} + \mathbf{F}$) performs best in Stage I, it is outperformed by ($\mathbf{x} + \mathbf{v} + \mathbf{F}$) in Stage II. 
This suggests that a useful teacher representation must not only distinguish intrinsic dynamics, but also contain cues that the video model can observe and match.
Displacement provides inter-frame motion cues that are directly visible in videos, while the deformation gradient captures fine-grained material responses.
Their combination balances dynamics discrimination and visual transfer, yielding the best Stage II performance. 
Accordingly, we used the full state in all experiments.

\subsection{Real-World Generalization}
\begin{figure}[!t]
	\centering
	\includegraphics[width=1\linewidth]{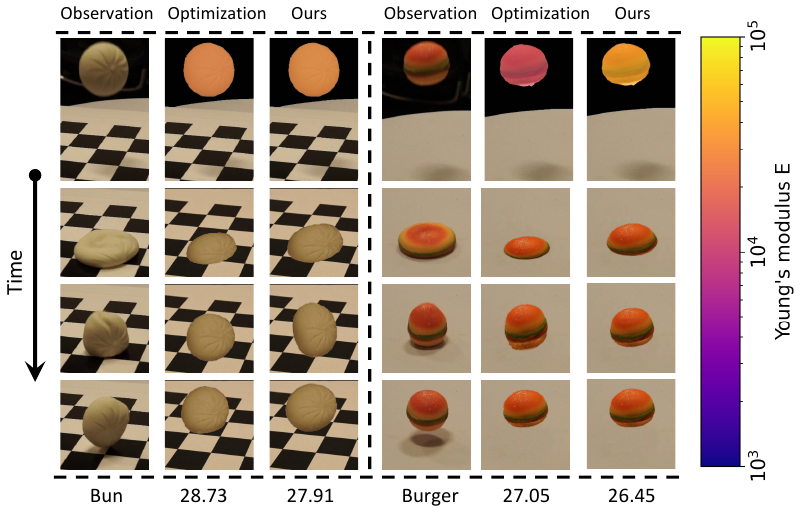}
	\caption{\textbf{Real-world dynamics reconstruction.} The mask color denotes the predicted Young’s modulus (E), following the right color bar. Without any fine-tuning, our method achieves performance comparable to per-scene optimization and reconstructs dynamics close to the observations.}
    \label{fig:real world}
\end{figure}

We further test whether a model trained only on synthetic data can directly generalize to real-world scenes. We evaluate on two real scenes, Bun and Burger, from Spring-Gaus~\cite{springgaus}. Since the ground-truth material parameters are unknown, we rely on PSNR as an indirect measure of the accuracy of material parameter inference.
As shown in Fig.~\ref{fig:real world}, our model predicts material parameters and simulates the dynamics without any fine-tuning on real videos. In contrast, per-scene optimization iteratively adjusts the parameters for each test scene. Despite this asymmetric setting, our PSNR is only 0.82 dB and 0.60 dB lower on Bun and Burger, respectively, while recovering deformation processes close to the observations. These results show that the dynamics learned from synthetic data can transfer to real observations, providing initial evidence of the framework’s applicability to real-world scenes.

We attribute the remaining gap mainly to the mismatch between the simulator and real-world physics. Our current simulator uses a limited set of idealized constitutive models~\cite{continuum}, while real objects may also exhibit damping, viscoelasticity, material heterogeneity, and complex contact behavior. Per-scene optimization can repeatedly adjust the available parameters for each video, partly compensating for these unmodeled effects and producing simulations closer to the observations. However, the optimized parameters may not reflect the object’s true physical properties. In contrast, our feed-forward model predicts directly from the synthetic training distribution without any additional per-scene refinement. Incorporating richer constitutive models and more accurate physical modeling is therefore an important direction for improving real-world generalization.

\section{Conclusion}
We presented~{\em Analytic Dynamics}, a feed-forward framework for efficiently inferring constitutive material models and parameters from monocular videos. Our key idea is to exploit privileged particle states available in simulation to learn a physics-grounded dynamics representation and then distill it into a video model. By providing the visual branch with a learning target derived from direct physical responses, our framework encourages video representations to capture dynamics-relevant patterns beyond object-specific visual correlations.
To enable systematic training and evaluation, we further developed an MPM-based data generation pipeline and benchmark that pair monocular mask videos with tracked physical-state trajectories and material annotations.
Extensive experiments demonstrate accurate material inference, strong generalization to unseen objects, and effective transfer to real-world observations, with an inference time of only 3.83 ms. Nevertheless, the current framework is trained on a limited set of idealized constitutive models. Extending it to richer material behaviors and interactions is a promising direction for future work.


\bibliography{aaai2027}

\clearpage

\appendix
\section*{Overview}
This appendix provides: 
i) additional details on dataset generation implementation; 
ii) extended experimental results and analyses; 
iii) the details of the MPM algorithm;
iv) the prompts used for vision-language baselines.

\section{Additional Dataset and Benchmark Details}
\subsection{Simulation and Rendering Settings}

We use the Material Point Method (MPM)~\cite{jiang2016mpm} to generate dynamics trajectories. Each normalized mesh is first converted into volumetric particles at a sampling resolution of $20$. During volume sampling, we establish correspondences between the sampled particles and the mesh vertices, allowing the simulated particle motion to be transferred back to the surface mesh for rendering. The sampled particles are then translated into the unit simulation domain $[0,1]^3$.

The simulation domain is discretized using an MPM grid of resolution $32$. We use a time step of $5\times10^{-4}$ and apply gravity $\mathbf{g}=9.8 m/s^2$ as the external excitation. All instances share the same ground boundary and contact settings. Each trajectory is simulated for 1,200 steps. We record the particle positions $\mathbf{x}$, displacements $\mathbf{v}$, and deformation gradients $\mathbf{F}$ throughout the simulation.


We render one frame every 50 simulation steps, including the initial state, resulting in 25 frames per sequence. All observations are rendered as monocular mask videos at a resolution of $512\times512$. During training, we randomly sample four views for each dynamics instance to reduce viewpoint bias. These views share the same underlying trajectory, physical states, and material annotations. During evaluation, we randomly select one view as input. The complete data-generation procedure is summarized in Algorithm~\ref{alg:data_generation}.

\subsection{Object Geometries and Data Splits}

Our benchmark contains 60 object geometries, including 25 synthetic shapes and 35 everyday objects selected from OmniObject3D~\cite{omniobject3d}. The synthetic set consists of basic primitives and manually constructed shapes, while the OmniObject3D objects provide more diverse and geometrically complex structures. We consider three material families: \texttt{jelly}, \texttt{plasticine}, and \texttt{sand}. Their constitutive formulations, varied parameters, and sampling ranges are summarized in Tab.~\ref{tab:constitutive_models}. 
Before particle sampling, each mesh is centered and uniformly scaled to fit within the normalized space $[-0.5,0.5]^3$.

The training set contains 30 geometries, comprising 10
synthetic shapes and 20 OmniObject3D objects. 
For each
geometry and each of the three material families, we sample
100 material-parameter configurations. 
Young's modulus and
yield stress are sampled uniformly in the logarithmic domain,
while Poisson's ratio and friction angle are sampled uniformly
in their original domains. 
This produces
$30 \times 3 \times 100 = 9{,}000$ training instances in total.

We construct two evaluation splits. The \emph{seen-object} split uses the same 30 geometries as the training set but contains newly sampled material parameters that are not used during training. We sample 10 parameter configurations for each geometry and each material family, resulting in $30 \times 3 \times 10 = 900$ instances.

The \emph{unseen-object} split contains the remaining 30 geometries, comprising 15 synthetic shapes and 15 OmniObject3D objects. None of these geometries appears during training. We again sample 10 parameter configurations for each geometry and each material family, resulting in  $30 \times 3 \times 10 = 900$ instances.

\begin{algorithm}[t]
\caption{Paired Dynamics Data Generation}
\label{alg:data_generation}
\begin{algorithmic}[1]
\Require Object mesh $\mathcal{M}$, material model $\Phi$,
parameters $\boldsymbol{\theta}$, and simulation configuration $\Omega$
\Ensure Physical trajectory $\mathcal{S}$ and multi-view videos $\mathcal{V}$

\State $\mathbf{x}^{0} \gets \Call{SampleParticles}{\mathcal{M}}$
\State $\mathbf{W} \gets \Call{BindMesh}{\mathcal{M},\mathbf{x}^{0}}$
\State $(\mathbf{v}^{0},\mathbf{F}^{0}) \gets \Call{InitializeState}{}$
\State $\mathcal{S},\mathcal{V} \gets \emptyset$

\For{$t=1$ to $T$}
    \State $(\mathbf{x}^{t},\mathbf{v}^{t},\mathbf{F}^{t})
    \gets \Call{MPMStep}{
    \mathbf{x}^{t-1},\mathbf{v}^{t-1},\mathbf{F}^{t-1},
    \Phi,\boldsymbol{\theta},\Omega}$

    \If{$t$ is an output frame}
        \State $\mathcal{S} \gets
        \mathcal{S}\cup
        \{(\mathbf{x}^{t},\mathbf{v}^{t},\mathbf{F}^{t})\}$
        \State $\mathbf{V}^{t}
        \gets \Call{UpdateMesh}{\mathcal{M},\mathbf{x}^{t},\mathbf{W}}$
        \State $\mathcal{V}
        \gets \mathcal{V}\cup
        \Call{RenderMultiView}{\mathbf{V}^{t}}$
    \EndIf
\EndFor

\State \Return $\mathcal{S},\mathcal{V}$
\end{algorithmic}
\end{algorithm}

\section{Additional Experimental Results}

\begin{table*}[t]
\centering
\begin{tabular}{c|ccc|cccc|cc}
\toprule
\multirow{2}{*}{Method} 
& \multicolumn{3}{c|}{Jelly}
& \multicolumn{4}{c|}{Plasticine}
& \multicolumn{2}{c}{Sand} \\

\cmidrule(lr){2-4}
\cmidrule(lr){5-8}
\cmidrule(lr){9-10}

& $Acc.\uparrow$
& $E\downarrow$
& $\nu\downarrow$

& $Acc.\uparrow$
& $E\downarrow$
& $\nu\downarrow$
& $Yield\downarrow$

& $Acc.\uparrow$
& $\phi\downarrow$ \\

\midrule

Per-scene Optimization
& - & 0.184 & 0.089 & - & 0.501 & 0.092 & 0.069 & - & 3.086
\\

\midrule




Ours (w/o alignment)
& 1.000 & 0.058 & 0.017 & 1.000 & 0.105 & 0.025 & \textbf{0.032} & 1.000 & \textbf{0.203}
\\

Ours (alignment only)
& 1.000 & 0.073 & 0.009 & 1.000 & 0.071 & 0.019 & 0.049 & 1.000 & 0.728
\\

Ours (Full)
& \textbf{1.000} & \textbf{0.053} & \textbf{0.008} & \textbf{1.000} & \textbf{0.062} & \textbf{0.018} & {0.035} & \textbf{1.000} & 0.327
\\

\bottomrule
\end{tabular}
\caption{
\textbf{Quantitative comparison on seen geometries with newly sampled material parameters.}
The test split uses the same 30 geometries as training but contains disjoint material-parameter configurations.
We report material classification accuracy ($\uparrow$) and material-specific parameter errors ($\downarrow$).
Young's modulus $E$ and yield stress $\sigma_y$ are evaluated using log-MAE, while Poisson's ratio $\nu$ and friction angle $\phi$ are evaluated using MAE.
}
\label{tab:seen_comparison}
\end{table*}

\subsection{Evaluation on Seen Objects}
We further isolate material-parameter inference from geometric generalization.
Specifically, we evaluate on the same 30 geometries used for training, while sampling new material-parameter configurations that are not observed during training.
As shown in Table~\ref{tab:seen_comparison}, all variants achieve nearly perfect material classification, indicating that material families can be reliably distinguished when the geometry is familiar.
Our full model provides the best overall regression performance, achieving the lowest errors for $E$ and $\nu$ on both jelly and plasticine, while substantially outperforming per-scene optimization across all three material families.

The gains from alignment are smaller than those observed on unseen objects, and the unaligned model remains slightly better for plastic yield stress and sand friction angle.
This is expected: when test geometries have already been observed during training, direct supervision can partially exploit geometry-correlated cues.
Under cross-shape generalization, however, these shortcuts become unreliable, making the shape-independent dynamics structure introduced by alignment more important.
These results show that our method preserves strong parameter interpolation on seen geometries, while it also improves generalization to unseen objects.

\begin{figure}[!t]
	\centering
	\includegraphics[width=1\linewidth]{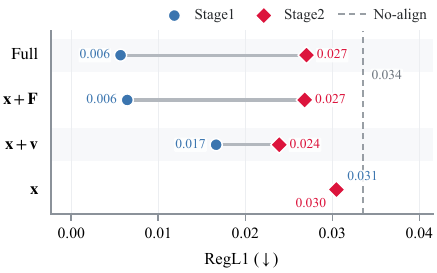}
	\caption{
    \textbf{Ablation of privileged physical states on seen objects with newly sampled material parameters.}
    We compare position $x$, displacement $v$, and deformation gradient $F$ as inputs to the Stage I teacher, and report the RegL1 ($\downarrow$) errors of both the teacher and its corresponding Stage II student. Gray lines connect each teacher--student pair, while the dashed line denotes the Stage II baseline without distillation.
    }
    \label{fig:state_ablation_seen}
\end{figure}

\subsection{Privileged-State Ablation on Seen Objects}
The main ablation evaluates privileged physical states on unseen geometries, where the full state achieves the best Stage-II performance.
To determine whether this advantage mainly arises from cross-geometry generalization, we repeat the same ablation on seen geometries with newly sampled material parameters.

As shown in Fig.~\ref{fig:state_ablation_seen}, all distilled students outperform the Stage-II baseline without alignment, confirming that privileged physical states provide a more informative learning target than material supervision alone.
For Stage I, adding the deformation gradient reduces RegL1 from $0.031$ with position alone to $0.006$, demonstrating its strong dynamics discriminability.
For Stage II, however, $\mathbf{x}+\mathbf{v}$ achieves the lowest error of $0.024$, slightly outperforming $\mathbf{x}+\mathbf{F}$ and the full state, both at $0.027$.
When geometry is familiar, displacement provides directly observable motion cues that the video model can readily match, reducing the benefit of additional deformation information.
In contrast, the full state performs best on unseen geometries, where combining visible motion with fine-grained deformation responses is more important for reducing shape dependence.
These results further show that privileged-state design must balance dynamics discriminability with visual transferability, and that the full state primarily benefits cross-shape generalization.

\begin{table*}[t]
\centering
\small
\renewcommand{\arraystretch}{1.12}

\begin{tabular*}{\linewidth}{
@{\extracolsep{\fill}}
l c|ccc|cccc|cc|c
@{}
}
\toprule
\multirow{2}{*}{Variant}
& \multirow{2}{*}{Frames}
& \multicolumn{3}{c|}{Jelly}
& \multicolumn{4}{c|}{Plasticine}
& \multicolumn{2}{c|}{Sand}
& \multirow{2}{*}{\shortstack{Overall\\RegL1$\downarrow$}}
\\

\cmidrule(lr){3-5}
\cmidrule(lr){6-9}
\cmidrule(lr){10-11}

&
& $Acc.\uparrow$
& $E\downarrow$
& $\nu\downarrow$
& $Acc.\uparrow$
& $E\downarrow$
& $\nu\downarrow$
& $Yield\downarrow$
& $Acc.\uparrow$
& $\phi\downarrow$
&
\\
\midrule

\multirow{5}{*}{Ours (w/o alignment)}
& \cellcolor{gray!12}25$^{\dagger}$
& \cellcolor{gray!12}0.997
& \cellcolor{gray!12}0.265
& \cellcolor{gray!12}0.043
& \cellcolor{gray!12}1.000
& \cellcolor{gray!12}0.223
& \cellcolor{gray!12}0.024
& \cellcolor{gray!12}0.115
& \cellcolor{gray!12}0.943
& \cellcolor{gray!12}1.991
& \cellcolor{gray!12}0.082
\\

& 24
& 0.997 & 0.286 & 0.042
& 1.000 & 0.230 & 0.024 & 0.114
& 0.943 & 2.004
& 0.082
\\

& 22
& 1.000 & 0.451 & 0.041
& 1.000 & 0.325 & 0.024 & 0.195
& 0.963 & 3.548
& 0.109
\\

& 18
& 1.000 & 0.776 & 0.057
& 0.993 & 0.792 & 0.025 & 0.657
& 0.867 & 7.217
& 0.185
\\

& 15
& 0.940 & 0.712 & 0.063
& 0.870 & 0.936 & 0.042 & 1.220
& 0.037 & 8.136
& 0.224
\\

\midrule

\multirow{5}{*}{Ours (Full)}
& \cellcolor{gray!12}25$^{\dagger}$
& \cellcolor{gray!12}1.000
& \cellcolor{gray!12}0.175
& \cellcolor{gray!12}0.037
& \cellcolor{gray!12}1.000
& \cellcolor{gray!12}0.141
& \cellcolor{gray!12}0.021
& \cellcolor{gray!12}0.101
& \cellcolor{gray!12}0.970
& \cellcolor{gray!12}1.777
& \cellcolor{gray!12}0.067
\\

& 24
& 1.000 & 0.180 & 0.035
& 1.000 & 0.146 & 0.021 & 0.099
& 0.970 & 1.803
& 0.068
\\

& 22
& 1.000 & 0.282 & 0.038
& 0.997 & 0.150 & 0.022 & 0.115
& 0.987 & 2.380
& 0.080
\\

& 18
& 0.993 & 0.519 & 0.047
& 0.973 & 0.296 & 0.025 & 0.283
& 0.963 & 3.550
& 0.117
\\

& 15
& 0.787 & 0.678 & 0.047
& 0.940 & 0.381 & 0.027 & 0.243
& 0.550 & 4.663
& 0.138
\\

\bottomrule
\end{tabular*}

\caption{
\textbf{Temporal OOD generalization on unseen objects under varying observation lengths.}
We evaluate the models using different numbers of observed frames
and report material classification accuracy ($\uparrow$),
material-specific parameter errors ($\downarrow$), and the overall
RegL1 error ($\downarrow$).
The 25-frame setting, marked by $^{\dagger}$ and shaded in gray,
is used during training, while the remaining observation lengths
are unseen during training.
}
\label{tab:temporal_ood}
\end{table*}

\begin{figure*}[!t]
	\centering
	\includegraphics[width=1\linewidth]{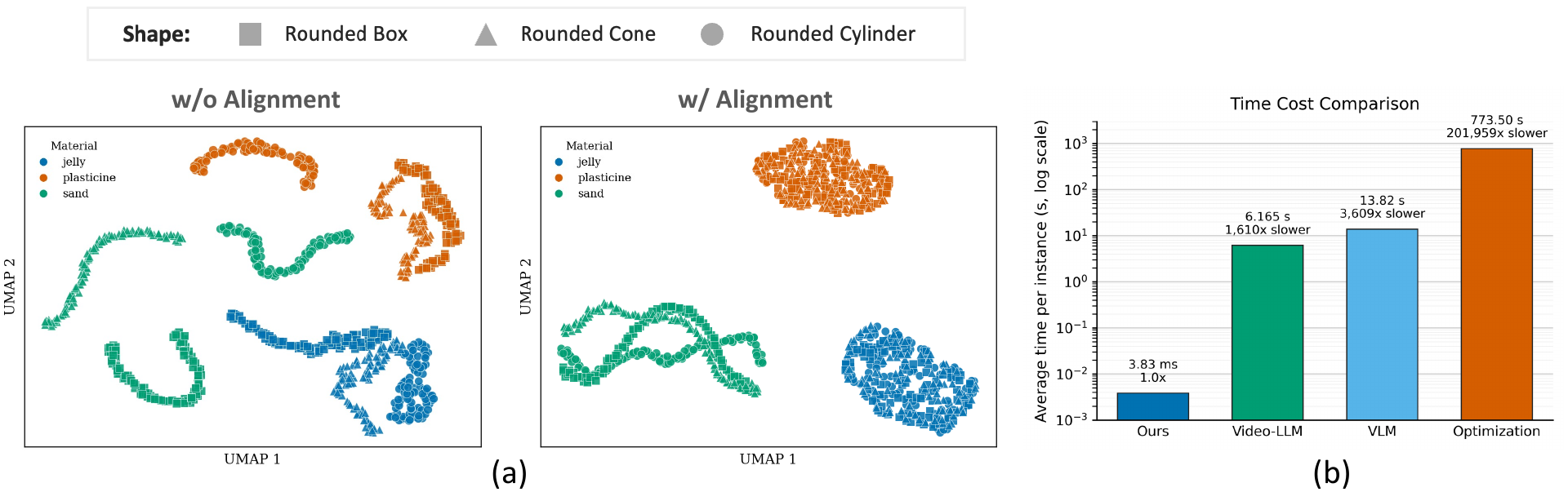}
    \caption{
    (a) UMAP visualizations of Stage-II representations without and with alignment. 
    Colors denote material families, while markers denote object geometries.
    Without alignment, samples form geometry-dependent branches; alignment brings different geometries of the same material closer, producing more compact and material-discriminative clusters.
    (b) Average inference time per instance on a logarithmic scale. Our feed-forward method requires only $3.83$\, ms, achieving orders-of-magnitude faster inference than Video-LLM, VLM, and per-scene optimization while maintaining accurate dynamics prediction.
    }
    \label{fig:alignment_efficiency}
\end{figure*}

\subsection{Temporal OOD Generalization on Unseen Objects}
We further examine whether the model remains reliable when only shorter, unseen temporal observations are available.
Keeping the unseen-object split fixed, we evaluate the models trained with 25-frame videos using the first $K\in\{24,22,18,15\}$ consecutive frames, and directly feed the resulting variable-length sequence to the video encoder. 
We compare the full model with its unaligned counterpart.

As shown in Table~\ref{tab:temporal_ood}, performance remains nearly unchanged with 24 frames but gradually degrades as the observation becomes shorter, since less motion and deformation evidence is available for dynamics inference.
The full model consistently outperforms the unaligned baseline, with a larger advantage under shorter observations.
At 15 frames, it reduces the overall RegL1 from $0.224$ to $0.138$.
This suggests that physics-grounded alignment helps the video model extract more stable dynamics cues from incomplete temporal evidence.
The remaining degradation under very short observations also indicates that sufficiently informative motion is still necessary for accurate material inference.

\subsection{Effect of Alignment and Inference Efficiency}
Figure~\ref{fig:alignment_efficiency}(a) provides additional visual evidence for the central finding in the main paper: physics-grounded alignment reduces the dependence of visual representations on object geometry. Without alignment, samples of the same material remain separated by geometry, forming shape-specific branches. After alignment, representations from different geometries cluster more compactly according to material family. This complementary visualization further confirms that distillation shifts the organization of the visual space from geometric cues toward intrinsic dynamics, consistent with the improved Shape-OOD performance reported in the main experiments.

Figure~\ref{fig:alignment_efficiency}(b) further compares the average inference time per instance. Our method requires only $3.83$ ms, achieving
orders-of-magnitude faster inference than VLM, Video-LLM, and per-scene optimization. This efficiency comes from predicting the material model and parameters in a single forward pass, without iterative simulation or test-time optimization. Together, these results show that our method learns a more dynamics-centered visual representation while retaining the efficiency of feed-forward inference.

\begin{table*}[t]
\centering
\small
\begin{tabular}{llll}
\toprule
Material
& Elastic model
& Plastic model
& Sampled parameters and ranges \\
\midrule
\texttt{Jelly}
& Corotated
& Identity
& $E \in [1.5\times10^{4},\,1.5\times10^{6}]$,
  $\nu \in [0.20,\,0.40]$ \\

\texttt{Plasticine}
& Sigma
& von Mises
& $E \in [5\times10^{4},\,5\times10^{5}]$,
  $\nu \in [0.35,\,0.48]$,
  $\sigma_y \in [500,\,25000]$ \\

\texttt{Sand}
& Sigma
& Drucker--Prager
& $E = 1 \times 10^5$,
  $\nu = 0.30$,
  $\phi_f \in [10^{\circ},\,45^{\circ}]$ \\
\bottomrule
\end{tabular}
\caption{
Constitutive models and parameter ranges used for each material family.
The selected parameter ranges span a broad spectrum of mechanical behaviors commonly observed within these material families. Here, $E$ and $\nu$ denote Young's modulus and Poisson's ratio,
respectively; $\phi_f$ denotes the friction angle; and
$\sigma_y$ is the yield stress.
For \texttt{sand}, only $\phi_f$ is varied because it is the dominant
parameter governing the observed frictional dynamics, while the remaining
elastic parameters are fixed.
}
\label{tab:constitutive_models}
\end{table*}

\section{Constitutive Material Families and Parameters}
Within the MPM framework, each material family is defined by an
elastic constitutive law and a plastic projection.
Given a trial deformation gradient $\mathbf{F}^{\mathrm{trial}}$,
the plastic model $\varphi_P$ first projects it onto the admissible elastic domain,
after which the elastic model $\varphi_E$ computes the Kirchhoff stress $\boldsymbol{\tau}$:
\begin{equation}
\mathbf{F}
=
\varphi_P\!\left(\mathbf{F}^{\mathrm{trial}}\right),
\qquad
\boldsymbol{\tau}
=
\varphi_E(\mathbf{F}).
\end{equation}

For a deformation gradient
$\mathbf{F}=\mathbf{U}\boldsymbol{\Sigma}\mathbf{V}^{T}$,
we define
\begin{equation}
J=\det(\mathbf{F}),
\qquad
\boldsymbol{\epsilon}=\log(\boldsymbol{\Sigma}),
\qquad
\hat{\boldsymbol{\epsilon}}
=
\boldsymbol{\epsilon}
-
\frac{\mathrm{tr}(\boldsymbol{\epsilon})}{d}\mathbf{I},
\end{equation}
where $d=3$ is the spatial dimension.
The Lamé parameters are computed from Young's modulus $E$ and
Poisson's ratio $\nu$ as
\begin{equation}
\mu=\frac{E}{2(1+\nu)},
\qquad
\lambda=\frac{E\nu}{(1+\nu)(1-2\nu)}.
\end{equation}

\subsection{Jelly Material}

We model \texttt{jelly} using
\texttt{CorotatedElasticity} together with
\texttt{IdentityPlasticity}.
This combination describes a purely elastic material without
irreversible deformation.

\paragraph{Identity plasticity.}
The identity projection leaves the trial deformation gradient unchanged:
\begin{equation}
\mathbf{F}
=
\mathbf{F}^{\mathrm{trial}}.
\end{equation}

\paragraph{Corotated elasticity.}
Given the polar rotation
\begin{equation}
\mathbf{R}=\mathbf{U}\mathbf{V}^{T},
\end{equation}
the Kirchhoff stress is
\begin{equation}
\boldsymbol{\tau}
=
2\mu
\left(\mathbf{F}-\mathbf{R}\right)\mathbf{F}^{T}
+
\lambda J(J-1)\mathbf{I}.
\label{eq:corotated_elasticity}
\end{equation}

The mechanical behavior of \texttt{jelly} is therefore controlled by
Young's modulus $E$ and Poisson's ratio $\nu$.
Young's modulus determines material stiffness, while Poisson's ratio
controls volumetric deformation.

\subsection{Plasticine Material}

We model \texttt{plasticine} using
\texttt{SigmaElasticity} together with
\texttt{VonMisesPlasticity}.
The elastic model describes the recoverable response, while the
von Mises projection introduces irreversible deformation once the
yield limit is exceeded.

\paragraph{Von Mises plasticity.}
Let
\begin{equation}
\mathbf{F}^{\mathrm{trial}}
=
\mathbf{U}\boldsymbol{\Sigma}\mathbf{V}^{T},
\qquad
\boldsymbol{\epsilon}
=
\log(\boldsymbol{\Sigma}).
\end{equation}
The plastic multiplier is defined as
\begin{equation}
\delta\gamma
=
\left\|
\hat{\boldsymbol{\epsilon}}
\right\|
-
\frac{\sigma_y}{2\mu},
\end{equation}
where $\sigma_y$ is the yield stress.
The corrected singular values are
\begin{equation}
\mathcal{Z}_{\mathrm{VM}}(\boldsymbol{\Sigma})
=
\begin{cases}
\boldsymbol{\Sigma},
& \delta\gamma \leq 0, \\[5pt]
\displaystyle
\exp\!\left(
\boldsymbol{\epsilon}
-
\delta\gamma
\frac{\hat{\boldsymbol{\epsilon}}}
{\|\hat{\boldsymbol{\epsilon}}\|}
\right),
& \delta\gamma > 0.
\end{cases}
\end{equation}
The corrected deformation gradient is then
\begin{equation}
\mathbf{F}
=
\mathbf{U}
\mathcal{Z}_{\mathrm{VM}}(\boldsymbol{\Sigma})
\mathbf{V}^{T}.
\label{eq:von_mises_plasticity}
\end{equation}

\paragraph{Sigma elasticity.}
For the corrected deformation gradient
$\mathbf{F}=\mathbf{U}\boldsymbol{\Sigma}\mathbf{V}^{T}$,
the Kirchhoff stress is computed in the logarithmic principal-strain
space:
\begin{equation}
\boldsymbol{\tau}
=
\mathbf{U}
\left(
2\mu\boldsymbol{\epsilon}
+
\lambda\,
\mathrm{tr}(\boldsymbol{\epsilon})\mathbf{I}
\right)
\mathbf{U}^{T}.
\label{eq:sigma_elasticity}
\end{equation}

The behavior of \texttt{plasticine} is controlled by
Young's modulus $E$, Poisson's ratio $\nu$, and yield stress $\sigma_y$.
The first two parameters determine its elastic response, while
$\sigma_y$ determines the onset of permanent deformation.

\subsection{Sand Material}

We model \texttt{sand} using
\texttt{SigmaElasticity} together with
\texttt{DruckerPragerPlasticity}.
The elastic response follows Eq.~\eqref{eq:sigma_elasticity}, while
the Drucker--Prager projection models pressure-dependent yielding and
frictional flow.

\paragraph{Drucker--Prager plasticity.}
Let
\begin{equation}
\mathbf{F}^{\mathrm{trial}}
=
\mathbf{U}\boldsymbol{\Sigma}\mathbf{V}^{T},
\qquad
\boldsymbol{\epsilon}
=
\log(\boldsymbol{\Sigma}).
\end{equation}
We define
\begin{equation}
\alpha
=
\sqrt{\frac{2}{3}}\,
\frac{2\sin\phi_f}{3-\sin\phi_f},
\end{equation}
where $\phi_f$ is the friction angle.
The plastic multiplier is
\begin{equation}
\delta\gamma
=
\left\|
\hat{\boldsymbol{\epsilon}}
\right\|
+
\alpha
\frac{
(d\lambda+2\mu)\,
\mathrm{tr}(\boldsymbol{\epsilon})
}{
2\mu
}.
\end{equation}

The corrected singular values are defined as
\begin{equation}
\mathcal{Z}_{\mathrm{DP}}(\boldsymbol{\Sigma})
=
\begin{cases}
\mathbf{I},
&
\mathrm{tr}(\boldsymbol{\epsilon})>0,
\\[5pt]
\boldsymbol{\Sigma},
&
\delta\gamma\leq 0
\ \text{and}\
\mathrm{tr}(\boldsymbol{\epsilon})\leq 0,
\\[5pt]
\displaystyle
\exp\!\left(
\boldsymbol{\epsilon}
-
\delta\gamma
\frac{\hat{\boldsymbol{\epsilon}}}
{\|\hat{\boldsymbol{\epsilon}}\|}
\right),
&
\text{otherwise}.
\end{cases}
\end{equation}
The corrected deformation gradient is
\begin{equation}
\mathbf{F}
=
\mathbf{U}
\mathcal{Z}_{\mathrm{DP}}(\boldsymbol{\Sigma})
\mathbf{V}^{T}.
\label{eq:drucker_prager_plasticity}
\end{equation}

The friction angle $\phi_f$ controls the shear resistance and flow
behavior of \texttt{sand}.
In our benchmark, we vary only $\phi_f$, which is the dominant factor
governing the observed sand dynamics, while keeping $E$ and $\nu$
fixed.

\section{Material Point Method Formulation}
\label{appendix:MPM}

In continuum mechanics~\cite{continuum}, material motion is governed by the conservation of mass and linear momentum:
\begin{equation}
\frac{D\rho}{Dt}
+
\rho \nabla \cdot \mathbf{v}
=
0,
\qquad
\rho \frac{D\mathbf{v}}{Dt}
=
\nabla \cdot \boldsymbol{\sigma}
+
\rho \mathbf{g},
\end{equation}
where $\rho$ is the density, $\mathbf{v}$ is the velocity field, $\mathbf{g}$ is the gravitational acceleration, and $\boldsymbol{\sigma}$ is the Cauchy stress.
The stress is determined by the constitutive law, which characterizes the material response to deformation.

We solve these equations using an explicit Material Point Method (MPM) solver with affine particle-in-cell transfers~\cite{jiang2016mpm}.
MPM represents the material using Lagrangian particles and evaluates physical interactions on a background Eulerian grid.
Each particle $p$ stores its position $\mathbf{x}_p$, mass $m_p$, reference volume $V_p^0$, velocity $\mathbf{v}_p$, deformation gradient $\mathbf{F}_p$, and affine velocity matrix $\mathbf{C}_p$.
Each simulation step consists of particle-to-grid transfer, grid update, and grid-to-particle transfer.

Let
\begin{equation}
w_{ip}^{t}
=
N_i(\mathbf{x}_p^{t}),
\qquad
\nabla w_{ip}^{t}
=
\nabla N_i(\mathbf{x}_p^{t}),
\end{equation}
denote the interpolation weight and its spatial gradient between particle $p$ and grid node $i$.
During particle-to-grid transfer, particle mass and momentum are accumulated on the grid:
\begin{align}
m_i^{t}
&=
\sum_p w_{ip}^{t} m_p,
\\
\mathbf{p}_i^{t}
&=
\sum_p
w_{ip}^{t} m_p
\left[
\mathbf{v}_p^{t}
+
\mathbf{C}_p^{t}
\left(
\mathbf{x}_i-\mathbf{x}_p^{t}
\right)
\right].
\end{align}
The internal force at each grid node is computed as
\begin{equation}
\mathbf{f}_i^{\mathrm{int},t}
=
-
\sum_p
V_p^0
\boldsymbol{\tau}_p^{t}
\nabla w_{ip}^{t},
\end{equation}
where $\boldsymbol{\tau}_p^{t}$ denotes the particle stress measure used by the discrete MPM solver.

For grid nodes with nonzero mass, the velocity is updated according to
\begin{align}
\mathbf{v}_i^{t}
&=
\frac{\mathbf{p}_i^{t}}{m_i^{t}},
\\
\widetilde{\mathbf{v}}_i^{t+1}
&=
\mathbf{v}_i^{t}
+
\Delta t
\left(
\frac{\mathbf{f}_i^{\mathrm{int},t}}{m_i^{t}}
+
\mathbf{g}
\right),
\\
\mathbf{v}_i^{t+1}
&=
\mathcal{B}
\left(
\widetilde{\mathbf{v}}_i^{t+1}
\right),
\end{align}
where $\mathcal{B}$ applies the prescribed boundary conditions and collision responses.

The updated grid velocities are then interpolated back to the particles:
\begin{align}
\mathbf{v}_p^{t+1}
&=
\sum_i
w_{ip}^{t}
\mathbf{v}_i^{t+1},
\\
\mathbf{x}_p^{t+1}
&=
\mathbf{x}_p^{t}
+
\Delta t\,\mathbf{v}_p^{t+1}.
\end{align}
Using quadratic B-spline interpolation, the affine velocity matrix is updated as
\begin{equation}
\mathbf{C}_p^{t+1}
=
\frac{4}{\Delta x^2}
\sum_i
w_{ip}^{t}
\mathbf{v}_i^{t+1}
\left(
\mathbf{x}_i-\mathbf{x}_p^{t}
\right)^{T},
\end{equation}
where $\Delta x$ is the grid spacing.

The trial deformation gradient is subsequently computed by
\begin{equation}
\mathbf{F}_p^{\mathrm{tr}}
=
\left(
\mathbf{I}
+
\Delta t\,\mathbf{C}_p^{t+1}
\right)
\mathbf{F}_p^{t}.
\end{equation}
We then apply the plastic projection and elastic stress mapping:
\begin{align}
\mathbf{F}_p^{t+1}
&=
\varphi_{P}
\left(
\mathbf{F}_p^{\mathrm{tr}};
\boldsymbol{\theta}_{P}
\right),
\\
\boldsymbol{\tau}_p^{t+1}
&=
\varphi_{E}
\left(
\mathbf{F}_p^{t+1};
\boldsymbol{\theta}_{E}
\right),
\end{align}
where $\varphi_P$ denotes the plastic projection, $\varphi_E$ denotes the elastic constitutive model, and $\boldsymbol{\theta}_{P}$ and $\boldsymbol{\theta}_{E}$ are their corresponding material parameters.
Here, $\mathbf{F}_p^{t+1}$ denotes the projected deformation gradient used for stress evaluation.
For purely elastic materials, $\varphi_P$ reduces to the identity mapping.

All material families share the same MPM update procedure and differ only in their elastic model, plastic model, and associated material parameters.

\section{Prompts for Vision-Language Baselines}
We use GPT-5.5 as the VLM baseline and Gemini 3.5 Flash
as the Video-LLM baseline. GPT-5.5 receives multi-view RGB
images of the target object, whereas Gemini 3.5 Flash receives
a monocular RGB video captured from a single view.

The system prompt for both models is set as:
\begin{lstlisting}[style=markdownstyle]
You are an expert assistant in continuum mechanics and Material Point Method (MPM) simulation.
\end{lstlisting}
For the VLM, the user prompt is designed for static spatial analysis:
\begin{lstlisting}[style=markdownstyle]
### TASK OVERVIEW
Given multi-view RGB images of a single 3D object, your task is to infer its physically plausible material model and material parameters for MPM simulation.

1. First, infer the most appropriate global material model for the whole object based on its visual appearance and expected physical behavior.
2. Then, infer the most likely continuous material parameter values required by the selected material model.

Treat the object as a single material body. Do not decompose it into parts.

### INPUTS
* Multi-view RGB images of the target object.

### Material Model Candidates
The material model must be selected from the following list:
- elastic
- plasticine
- sand

### Material Parameters
The required material parameters depend on the selected material model:

- elastic:
  - Young's Modulus
  - Poisson's Ratio

- plasticine:
  - Young's Modulus
  - Poisson's Ratio
  - yield stress

- sand:
  - friction angle
  
### Parameter meaning
Young's Modulus controls stiffness. A larger value means the object is harder to deform, while a smaller value means it is softer or more flexible.

Poisson's Ratio controls lateral expansion under compression. For most solid materials, it should be between 0 and 0.5.

Yield stress controls when a plastic material starts to undergo irreversible deformation. A larger value means the material resists plastic flow more strongly.

Friction angle controls the internal friction of granular material. A larger value means the material forms steeper piles and resists flow more strongly.

### REQUIRED OUTPUT (JSON)
You MUST output a JSON object containing:
{
  "material": {
    "material_model": "elastic | plasticine | sand",
    "material_parameters": {
      "...": value
    }
  },
  "reasoning": "Brief explanation of the visual cues and physical intuition used to select the material model and parameter values."
}

### Important Rules
1. Select exactly one material model.
2. Only output the parameters required by the selected material model.
3. Do not output parameters for unselected material models.
4. All continuous parameters must be output as single scalar values, not ranges.
5. The parameter values should be physically plausible for MPM simulation.
6. If the visual evidence is ambiguous, choose the most likely material model and output conservative parameter values.
7. Output only valid JSON. Do not include markdown, comments, or any extra text outside the JSON object.
\end{lstlisting}
For the Video-LLM, the user prompt is designed for temporal evolution analysis:
\begin{lstlisting}[style=markdownstyle]
### TASK OVERVIEW
Given a motion video of a single object, your task is to infer its physically plausible material model and material parameters for MPM simulation.

You should perform the inference in two steps:

1. First, infer the most appropriate global material model for the whole object based on its observed motion, deformation behavior, and visual appearance.
2. Then, infer the most likely continuous material parameter values required by the selected material model.

Treat the object as a single material body. Do not decompose it into parts.

The video contain object motion caused by gravity, collision, compression. You should use the dynamic cues in the video as the primary evidence, and use visual appearance as secondary evidence when the motion is ambiguous.

### INPUTS
* A motion video of the target object.

### MATERIAL MODEL CANDIDATES
The material model must be selected from the following list:
- elastic
- plasticine
- sand

### MATERIAL PARAMETERS
The required material parameters depend on the selected material model:

- elastic:
  - Young's Modulus
  - Poisson's Ratio

- plasticine:
  - Young's Modulus
  - Poisson's Ratio
  - yield stress

- sand:
  - friction angle

### PARAMETER MEANING
Young's Modulus controls stiffness. A larger value means the object is harder to deform, while a smaller value means it is softer or more flexible.

Poisson's Ratio controls lateral expansion under compression. For most solid materials, it should be between 0 and 0.5.

Yield stress controls when a plastic material starts to undergo irreversible deformation. A larger value means the material resists plastic flow more strongly.

Friction angle controls the internal friction of granular material. A larger value means the material forms steeper piles and resists flow more strongly.

### DYNAMIC CUES TO CONSIDER
Use the object's motion and deformation behavior in the video to guide your inference:

- If the object deforms under force but recovers its original shape, bounces, oscillates, or shows reversible deformation, it is likely elastic.
- If the object deforms and does not fully recover, keeps a dent, flattens, spreads, or undergoes permanent shape change, it is likely plasticine.
- If the object consists of many small grains, flows, piles up, collapses, avalanches, or forms a slope under gravity, it is likely sand.

For elastic materials:
- Larger Young's Modulus should be assigned when the object appears stiff, deforms only slightly, rebounds quickly, or preserves its shape.
- Smaller Young's Modulus should be assigned when the object appears soft, bends strongly, compresses visibly, or oscillates slowly.
- Poisson's Ratio should be higher for nearly incompressible soft materials such as rubber-like objects.

For plasticine materials:
- Larger Young's Modulus should be assigned when the object resists deformation before yielding.
- Smaller Young's Modulus should be assigned when the object is easily compressed or reshaped.
- Larger yield stress should be assigned when permanent deformation starts only under strong force.
- Smaller yield stress should be assigned when the object yields easily and retains deformation under weak force.

For sand materials:
- Larger friction angle should be assigned when the granular material forms a steep stable pile and resists flow.
- Smaller friction angle should be assigned when it flows easily, spreads widely, or forms a shallow pile.

### REQUIRED OUTPUT (JSON)
You MUST output a JSON object containing:
{
  "material": {
    "material_model": "elastic | plasticine | sand",
    "material_parameters": {
      "...": value
    }
  },
  "reasoning": "Brief explanation of the observed motion/deformation cues and physical intuition used to select the material model and parameter values."
}

### IMPORTANT RULES
1. Select exactly one material model.
2. Only output the parameters required by the selected material model.
3. Do not output parameters for unselected material models.
4. All continuous parameters must be output as single scalar values, not ranges.
5. The parameter values should be physically plausible for MPM simulation.
6. Use dynamic cues from the video as the primary evidence.
7. Use visual appearance only as secondary evidence when motion cues are insufficient or ambiguous.
8. If the video does not clearly show deformation or interaction, choose the most likely material model based on appearance and output conservative parameter values.
9. Output only valid JSON. Do not include markdown, comments, or any extra text outside the JSON object.
\end{lstlisting}


\end{document}